\documentclass[letterpaper]{article}

\usepackage{aaai19}
\usepackage{times}
\usepackage{helvet}
\usepackage{courier}
\usepackage{url}   
\usepackage{graphicx}
\usepackage[utf8]{inputenc} 
\usepackage{booktabs}       
\usepackage{amsfonts}       
\usepackage{nicefrac}       
\usepackage{microtype}      
\usepackage{amsmath}
\usepackage{algorithm}
\usepackage{algpseudocode}
\usepackage{mathtools}
\usepackage{enumitem}
\usepackage{amsthm}

\usepackage[hidelinks]{hyperref}
\usepackage{natbib}

\newcommand{\argmin}{\mathop{\mathrm{argmin}}}
\newcommand{\argmax}{\mathop{\mathrm{argmax}}}
\newcommand{\expect}{\mathop{\mathbb{E}}}
  
\graphicspath{ {img/} }
\title{Classification with Costly Features using Deep Reinforcement Learning}
\author{
  Jaromír Janisch \and Tomáš Pevný \and Viliam Lisý\\
  Artificial Intelligence Center, Department of Computer Science\\
  Faculty of Electrical Engineering, Czech Technical University in Prague\\
  \{jaromir.janisch, tomas.pevny, viliam.lisy\}@fel.cvut.cz \\
}

\begin{document}
\maketitle

\begin{abstract}
We study a classification problem where each feature can be acquired for a cost and the goal is to optimize a trade-off between the expected classification error and the feature cost. We revisit a former approach that has framed the problem as a sequential decision-making problem and solved it by Q-learning with a linear approximation, where individual actions are either requests for feature values or terminate the episode by providing a classification decision. On a set of eight problems, we demonstrate that by replacing the linear approximation with neural networks the approach becomes comparable to the state-of-the-art algorithms developed specifically for this problem. The approach is flexible, as it can be improved with any new reinforcement learning enhancement, it allows inclusion of pre-trained high-performance classifier, and unlike prior art, its performance is robust across all evaluated datasets.
\end{abstract}

\section{Introduction}

In real-world classification problems, one often has to deal with limited resources - time, money, computational power and many other. Medical practitioners strive to make a diagnosis for their patients with high accuracy. Yet at the same time, they want to minimize the amount of money spent on examinations, or the amount of time that all the procedures take. In the domain of computer security, network traffic is often analyzed by a human operator who queries different expensive data sources or cloud services and eventually decides whether the currently examined traffic is malicious or benign. In robotics, the agent may utilize several measurement devices to decide its current position. Here, each measurement has an associated cost in terms of the energy consumption. In all of these cases, the agent gathers a set of \textit{features}, but it is not desirable to have a static set that works on average. Instead, we want to optimize for a specific sample - the current patient, certain computer on a network or the immediate robot's location.

These real-world problems give rise to the problem of Classification with Costly Features (CwCF). In this setting, an algorithm has to classify a sample, but can only reveal its features at a defined cost. Each sample is treated independently, and for each sample the algorithm \emph{sequentially} selects features conditioning on values already revealed. Inherently, a different subset of features can be selected for different samples. The goal is to minimize the expected classification error, while also minimizing the expected incurred cost.

\begin{figure}[t]
  \centering
  \includegraphics[width=0.90\linewidth]{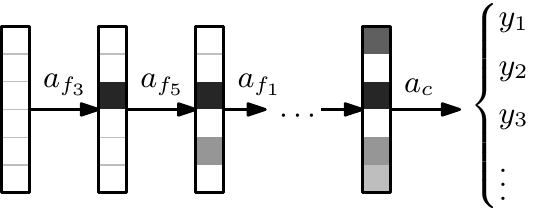}

  \caption{Illustrative sequential process of classification. The agent sequentially asks for different features (actions $a_{f}$) and finally performs a classification ($a_c$). The particular decisions are influenced by the observed values.}

  \label{fig:seq_process}
\end{figure}

In this paper, we extend the approach taken by \citet{dulac2011datum}, which proposed to formalize the problem as an Markov decision process (MDP) and solve it with linearly approximated Q-learning. In this formalization, each sample corresponds to an episode, where an agent sequentially decides whether to acquire another feature and which, or whether to already classify the sample (see Figure\;\ref{fig:seq_process}). At each step, the agent can base its decision on the values of the features acquired so far. For the actions requesting a feature, the agent receives a negative reward, equal to the feature cost. For the classification actions, the reward is based on whether the prediction is correct. \citeauthor{dulac2011datum} prove in their paper that the optimal solution to this MDP corresponds to the optimal solution of the original CwCF problem.

Since \citeyear{dulac2011datum}, we are not aware of any work improving upon the method of \citeauthor{dulac2011datum} In this paper, we take a fresh look at the method and show that simple replacement of the linear approximation with neural networks can outperform the most recent methods specifically designed for the CwCF problem. We take this approach further and implement several state-of-the-art techniques from Deep Reinforcement Learning (Deep RL), where we show that they improve performance, convergence speed and stability. We argue that our method is extensible in various ways and we implement two extensions: First we identify parts of the model that can be pre-trained in a fast and supervised way, which improves performance initially during the training. Second, we allow the model to selectively use an external High-Performance Classifier (HPC), trained separately with all features. In many real-world domains, there is an existing legacy, cost-agnostic model, which can be utilized. This approach is similar to \citep{nan2017adaptive}, but in our case it is a straightforward extension to the baseline model. We evaluate and compare the method on several two- and multi-class public datasets with number of features ranging from 8 to 784. We do not perform any thorough hyperparameter search for each dataset, yet our method robustly performs well on all of them, while often outperforming the prior-art algorithms. The source code is available at \url{https://github.com/jaromiru/cwcf}.

\section{Problem definition}
Let $(x,y) \in \mathcal{D}$ be a sample drawn from a data distribution $\mathcal{D}$. Vector $x \in \mathcal{X} \subseteq \mathbf{R}^n$ contains feature values, where $x_i$ is a value of feature $f_i \in \mathcal{F} = \{f_1, ..., f_n\}$, $n$ is the number of features, and $y \in \mathcal{Y}$ is a class. Let $c : \mathcal{F} \rightarrow \mathbf{R}$ be a function mapping a feature $f$ into its real-valued cost $c(f)$, and let $\lambda \in [0, 1]$ be a cost scaling factor.

The model for classifying one sample is a parametrized couple of functions $y_\theta: \mathcal{X} \rightarrow \mathcal{Y}$, $z_\theta: \mathcal{X} \rightarrow \wp(\mathcal{F})$, where $y_\theta$ classifies and $z_\theta$ returns the features used in the classification. The goal is to find such parameters $\theta$ that minimize the expected classification error along with $\lambda$ scaled expected feature cost. That is:
\begin{equation}\label{eq:problem}
\argmin_\theta \frac{1}{|\mathcal{D}|} \sum_{(x,y) \in \mathcal{D}} \left[ \ell(y_\theta(x), y) + \lambda \sum_{f \in z_\theta(x)} c(f) \right]
\end{equation}

We view the problem as a sequential decision-making problem, where at each step an agent selects a feature to view or makes a class prediction. We use standard reinforcement learning setting, where the agent explores its environment through actions and observes rewards and states. We represent this environment as a partially observable Markov decision process (POMDP), where each episode corresponds to a classification of one sample from a dataset. We use POMDP definition, because it allows a simple definition of the transition mechanics. However, our model solves a transformed MDP with stochastic transitions and rewards.

Let $\mathcal{S}$ be the state space, $\mathcal{A}$ set of actions and $r$, $t$ reward and transition functions. State $s = (x, y, \mathcal{\bar{F}}) \in \mathcal{S}$ represents a sample $(x,y)$ and currently selected set of features $\mathcal{\bar{F}}$. The agent receives only an observation $o = \{ (x_i,f_i) \mid \forall f_i \in \mathcal{\bar{F}} \}$, that is, the selected parts of $x$ without the label.

\begin{figure}[t]
  \centering
  \includegraphics[width=0.5\linewidth]{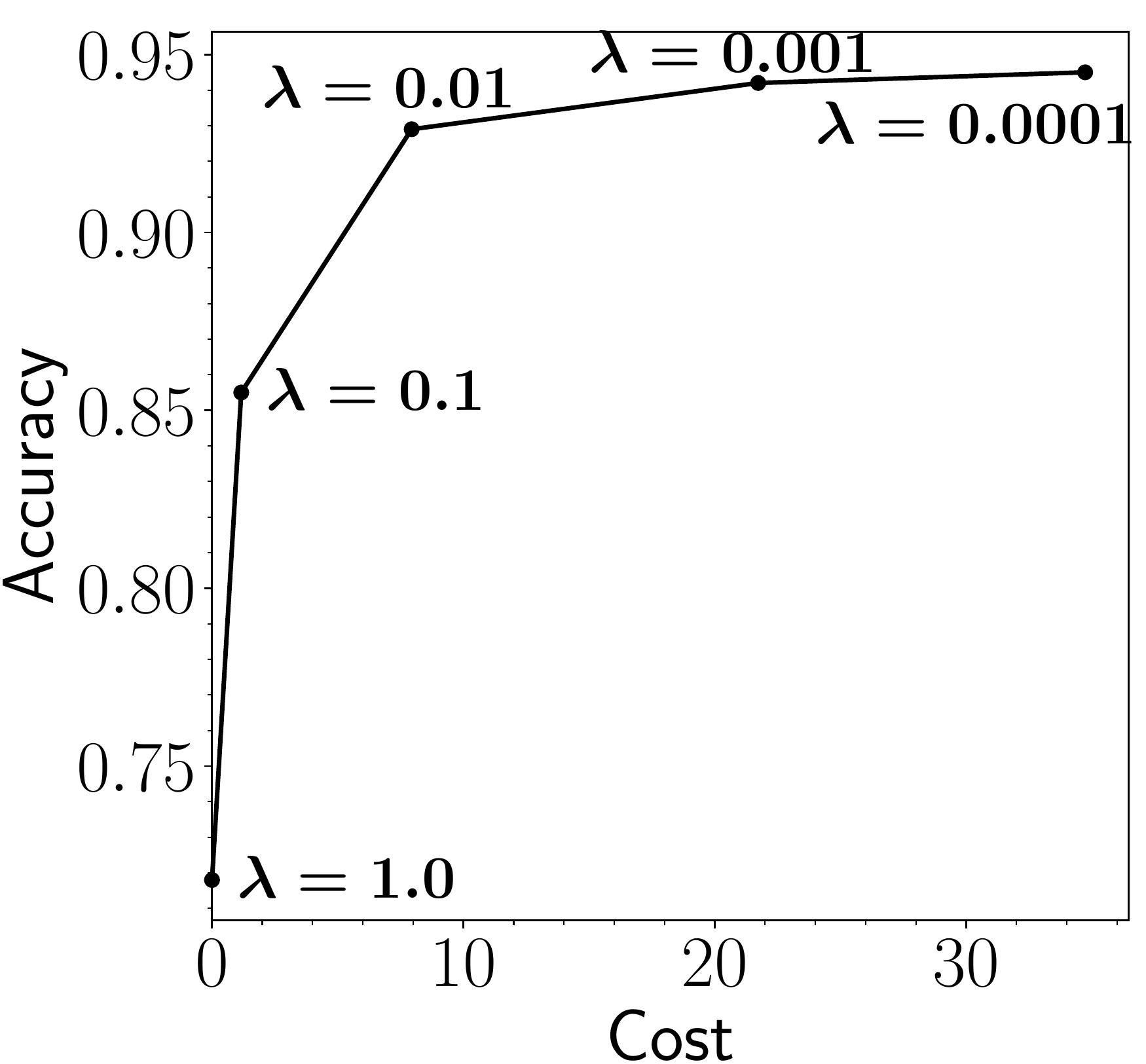}

  \caption{Different settings of $\lambda$ result in different cost-accuracy trade-off. Five different runs, miniboone dataset.}

  \label{fig:tradeoff}
\end{figure}

Action $a \in \mathcal{A} = \mathcal{A}_c \cup \mathcal{A}_f$ is one of the classification actions $\mathcal{A}_c = \mathcal{Y}$, or feature selecting actions $\mathcal{A}_f = \mathcal{F}$. Classification actions $\mathcal{A}_c$ terminate the episode and the agent receives a reward of $0$ in case of correct classification, else $-1$. Feature selecting actions $\mathcal{A}_f$ reveal the corresponding value $x_i$ and the agent receives a reward of $-\lambda c(f_i)$. The set of available feature selecting actions is limited to features not yet selected. Reward function $r: S \times \mathcal{A} \rightarrow \mathbf{R}$ is:
$$ r((x, y, \mathcal{\bar F}), a) = 
  \begin{dcases*}
    -\lambda c(f_i)       & if $a \in \mathcal{A}_f$, $a = f_i$\\
    0                   & if $a \in \mathcal{A}_c$ and $a = y$ \\
    -1                  & if $a \in \mathcal{A}_c$ and $a \neq y$ \\
  \end{dcases*}
$$
By altering the value of $\lambda$, one can make a trade-off between precision and average cost. Higher $\lambda$ forces the agent to prefer lower cost and shorter episodes over precision and vice versa. Further intuition can be gained by looking at Figure\;\ref{fig:tradeoff}.

The initial state does not contain any disclosed features, $s_0 = (x, y, \emptyset)$, and is drawn from the data distribution $\mathcal{D}$. The environment is deterministic with a transition function $t: \mathcal{S} \times \mathcal{A} \rightarrow \mathcal{S} \cup \mathcal{T}$, where $\mathcal{T}$ is the terminal state: 
$$ t((x, y, \mathcal{\bar{F}}), a) = 
  \begin{dcases*}
    \mathcal{T}  & if $a \in \mathcal{A}_c$\\
    (x, y, \mathcal{\bar{F}} \cup a)   & if $a \in \mathcal{A}_f$
  \end{dcases*}
$$
These properties make the environment inherently episodic, with a maximal length of an episode of $|\mathcal{F}|+1$. Finding the optimal policy $\pi_\theta$ that maximizes the expected reward in this environment is equivalent to solving eq. \eqref{eq:problem}, which was shown by \citet{dulac2011datum}.

\section{Background}
In Q-learning, one seeks the optimal function $Q^*$, representing the expected discounted reward for taking an action $a$ in a state $s$ and then following the optimal policy, and it satisfies the Bellman equation:
\begin{equation} \label{eq:beleq}
  Q^*(s, a) = \expect_{s' \sim t(s,a)} \left[r(s,a,s') + \gamma \max_{a'} Q^*(s', a') \right]
\end{equation}
where $r(s, a, s')$ is the received reward and $\gamma \leq 1$ is the discount factor. A neural network with parameters $\theta$ takes a state $s$ and outputs an estimate $Q^\theta(s,a)$, jointly for all actions $a$. It is optimized by minimizing MSE between the both sides of eq. \eqref{eq:beleq} for transitions $(s_t, a_t, r_t, s_{t+1})$ empirically experienced by an agent following a greedy policy $\pi_\theta(s) = \argmax_a Q^\theta(s, a)$. Formally, we are looking for parameters $\theta$ by iteratively minimizing the loss function $\ell_\theta$, for a batch of transitions $\mathcal{B}$:
\begin{equation} \label{eq:loss}
\ell_\theta(\mathcal{B}) = \frac{1}{|\mathcal{B}|} \sum_{(s_t,a_t,r_t,s_{t+1}) \in \mathcal{B}} (q_t - Q^\theta(s_t, a_t))^2
\end{equation}
where $q_t$ is regarded as a constant when differentiated, and is computed as:
\begin{equation} \label{eq:loss_q}
q_t = 
    \begin{dcases*}
      r_t                                             & if $s_{t+1} = \mathcal T$ \\
      r_t + \max_{a} \gamma Q^\theta(s_{t+1}, a)    & otherwise
    \end{dcases*}
\end{equation}
As the error decreases, the approximated function $Q^\theta$ converges to $Q^*$. However, this method proved to be unstable in practice \citep{mnih2015human}. In the following section, we briefly describe techniques used in this work that stabilize and speed-up the learning.

\textbf{Deep Q-learning} \citep{mnih2015human} includes a separate \textit{target network} with parameters $\phi$, which follow parameters $\theta$ with a delay. Here we use the method of \citet{lillicrap2015continuous}, where the weights are regularly updated with expression $\phi := (1-\rho) \phi + \rho \theta$, with some parameter $\rho$. The slowly changing estimate $Q^\phi$ is then used in $q_t$, when $s_{t+1} \neq \mathcal T$:
\begin{equation} \label{eq:loss_dqn}
q_t = r_t + \max_{a} \gamma Q^\phi(s_{t+1}, a)
\end{equation}

\textbf{Double Q-learning} \citep{van2016deep} is a technique to reduce bias induced by the \textit{max} in eq. \eqref{eq:loss_q}, by combining the two estimates $Q^\theta$ and $Q^\phi$ into a new formula for $q_t$, when $s_{t+1} \neq \mathcal T$:
\begin{equation}
q_t = r_t + \gamma Q^\phi(s_{t+1}, \argmax_{a} Q^\theta(s_{t+1}, a))
\end{equation}
In the expression, the maximizing action is taken from $Q^\theta$, but its value is estimated with the target network $Q^\phi$.

\textbf{Dueling Architecture} \citep{wang2016dueling} uses a decomposition of the Q-function into two separate value and advantage functions. The architecture of the network is altered so that the it outputs two estimates $V^\theta(s)$ and $A^\theta(s, a)$ for all actions $a$, which are then combined to a final output $Q^\theta(s, a) = V^\theta(s) + A^\theta(s, a) - \frac{1}{|\mathcal A|}\sum_{a'} A^\theta(s, a')$. When training, we take the gradient w.r.t. the final estimate $Q^\theta$. By incorporating baseline $V^\theta$ across different states, this technique accelerates and stabilizes training.

\textbf{Retrace} \citep{munos2016safe} is a method to efficiently utilize long traces of experience with truncated importance sampling. We store generated trajectories into an experience replay buffer \citep{lin1993reinforcement} and utilize whole episode returns by recursively expanding eq. \eqref{eq:beleq}. The stored trajectories are off the current policy and a correction is needed. For a sequence $(s_0, a_0, r_0, \dots, s_n, a_n, r_n, \mathcal T)$, we implement Retrace together with Double Q-learning by replacing $q_t$ with
\begin{equation} \label{eq:loss_q_full}
\begin{split}
q_t &= r_t + \gamma \expect_{a \sim \pi_\theta(s_t)} \left[ Q^\phi(s_{t+1}, a) \right] \; + \\
    &+ \gamma \bar\rho_{t+1} \left[q_{t+1} - Q^\phi(s_{t+1}, a_{t+1}) \right]
\end{split}
\end{equation}
Where we define $Q^\phi(\mathcal T, \cdot) = 0$ and $\bar\rho_t = \min(\frac{\pi(a_t|s_t)}{\mu(a_t|s_t)}, 1)$ is a truncated importance sampling between exploration policy $\mu$ that was used when the trajectory was sampled and the current policy $\pi$. 
We allow the policy $\pi_\theta$ to be stochastic -- at the beginning, it starts close to the sampling policy $\mu$ but becomes increasingly greedy as the training progresses. It prevents premature truncation in the eq. \eqref{eq:loss_q_full} and we observed faster convergence. Note that all $q_t$ values for a whole episode can be calculated in $\mathcal O(n)$ time. Further, it can be easily parallelized across all episodes.

\section{Method}
We first describe a Deep Q-learning (DQN) based method, and leave other extensions to the next section. 

\begin{figure}[t]
  \centering
  \includegraphics[width=0.40\linewidth]{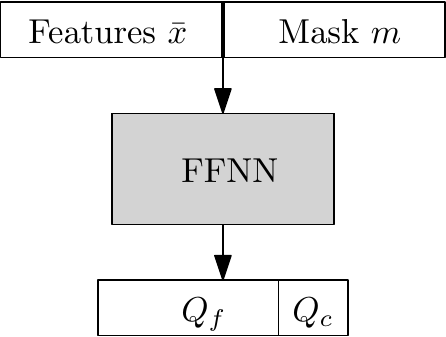}
  \caption{The architecture of the model. The input layer consists of the feature vector $\bar x$ concatenated with the binary mask $m$, followed by a feed-forward neural network (FFNN). Final fully connected layer jointly outputs Q-values for both classification and feature-selecting actions.}
  \label{fig:nn_arch}
\end{figure}

An observation $o$ is mapped into a tuple $(\bar x, m)$:
$$
\bar{x}_i =
  \begin{dcases*}
     x_i  &  if $f_i \in \mathcal{\bar{F}}$ \\
     0        &  otherwise
  \end{dcases*}, 
m_i =
  \begin{dcases*}
     1    &  if $f_i \in \mathcal{\bar{F}}$ \\
     0    &  otherwise
  \end{dcases*}
$$
Vector $\bar{x} \in \mathbf{R}^n$ is a masked vector of the original $x$. It contains values of $x$ which have been acquired and zeros for unknown values. 
Mask $m \in \{0, 1\}^n$ is a vector denoting whether a specific feature has been acquired, and it contains 1 at a position of acquired features, or 0. The combination of $\bar{x}$ and $m$ is required so that the model can differentiate between a feature not present and observed value of zero. Each dataset is normalized with its mean and standard deviation and because we replace unobserved values with zero, this corresponds to the mean-imputation of missing values.

The neural network is fed with concatenated vectors $\bar{x}$, $m$ and output Q-values jointly for all actions. The neural network has three fully connected hidden layers, each followed by the ReLu non-linearity, where the number of neurons in individual layers change depending on the used dataset. The overview is shown in Figure\;\ref{fig:nn_arch}.

A set of environments with samples randomly drawn from the dataset are simulated and the experienced trajectories are recorded into the experience replay buffer. After each action, a batch of transitions $\mathcal{B}$ is taken from the buffer and optimized upon with Adam \citep{kingma2014adam}, with eqs. (\ref{eq:loss}, \ref{eq:loss_dqn}). The gradient is normalized before back-propagation if its norm exceeds $1.0$. The target network is updated after each step. Overview of the algorithm and the environment simulation is in Algorithm \ref{alg:rl} and \;\ref{alg:env}.

Because all rewards are non-positive, the whole Q-function is also non-positive. We use this knowledge and clip the $q_t$ value so that it is at most $0$. We experimentally found that this approach avoids an initial explosion in predicted values and weights and speeds up and stabilizes the learning process. The definition of the reward function also results in optimistic initialization: Since the outputs of an untrained network are around zero, the model tends to overestimate the real Q-values, which has a positive effect on exploration.

The environment is episodic with a short average length of episodes, therefore we use undiscounted returns with $\gamma=1.0$. We use $\epsilon$-greedy policy that behaves greedily most of the time, but picks a random action with probability $\epsilon$. Exploration rate $\epsilon$ starts at a defined initial value and it is linearly decreased over time to its minimum value. 

\subsection{Extensions}

\subsubsection{Pre-training}
Classification actions $\mathcal A_c$ are terminal and their Q-values do not depend on any following states. Hence, these Q-values can be efficiently pretrained in a supervised way. Before starting the main method, we repeatedly sample different states by randomly picking a sample $x$ from the dataset and generating a mask $m$. The values $m_i$ follow the Bernoulli distribution with probability $p$. As we want to generate states with different amount of observed features, we randomly select $\sqrt[3]{p} \sim \mathcal U(0, 1)$ for different states. This way, the distribution of generated masks $m$ is shifted towards the initial state with no observed features. We use the generated states $s$ to pretrain the part of the network $Q^\theta(s, a)$, for classification actions $a \in \mathcal A_c$. Since the main algorithm starts with already accurate classification predictions, this technique has a positive effect on the initial speed of the training process.

\subsubsection{HPC}
Our second extension is the High-Performance Classifier. This can coincide with an expensive and (not-always) accurate oracle, that appears in real-world problems (e.g. the human operator in computer security), or a legacy cost-agnostic system already in place. We model these cases with a separately trained classifier, typically of a different type than neural networks (e.g. random forests or SVM). The extension is implemented by an addition of a separate action $a_\text{HPC}$, that corresponds to forwarding the current sample to the HPC. The cost for this action is the summed cost of all remaining features, i.e. $r(s, a_\text{HPC}) = -\lambda \sum_{f \in \mathcal{F} \setminus \mathcal{\bar F}} c(f)$. The action is terminal, $t(s,a_\text{HPC}) = \mathcal T$. The model learns to use this additional action, which has two main effects: (1) It improves performance for samples needing a large amount of features. (2) It offloads the complex samples to HPC, so the model can focus its capacity more effectively for the rest, improving overall performance. Note that HPC is an optional extension, but is likely to improve performance if the chosen classifier outperforms neural networks on the particular dataset.

\subsubsection{Deep RL extensions}
Since our method is based on a generic RL technique, any domain-independent improvement can be included into the algorithm. Here we decide to include the three described techniques: Double Q-learning, Dueling Architecture and Retrace. Compared to the DQN model, a few changes are necessary: 
The architecture is changed to Dueling and the optimization step is made on eqs. (\ref{eq:loss}, \ref{eq:loss_q_full}). The baseline agent uses one-step return in eq. \eqref{eq:loss_dqn}, but the full agent with Retrace updates the Q-values along the whole sequence with eq. \eqref{eq:loss_q_full}. In the latter case, we sample whole episodes instead of individual steps, but we keep the total number of steps in batches similar.  The target policy is set to be $\eta$-greedy, where $\eta$ is linearly decreased towards zero. This way, the target and exploration policy are close together at the beginning of the training, which avoids premature truncations in eq. \eqref{eq:loss_q_full} and speeds-up the training.

\begin{algorithm}[t]
\caption{Training}
\label{alg:rl}
\begin{algorithmic}
\State Randomly initialize parameters $\theta$ and $\phi$
\State Initialize environments $\mathcal{E}$ with $(x, y, \emptyset) \in (\mathcal{X}, \mathcal{Y}, \wp(\mathcal{F}))$
\State Initialize replay buffer $\mathcal{M}$ with a random agent

\While{not converged}
  \ForAll {$e \in \mathcal{E}$}
    \State Simulate one step with $\epsilon$-greedy policy $\pi_\theta$:
      $$a = \pi_\theta(s);\quad s', r = \Call{step}{e, a}$$
    \State Add transition $(s, a, r, s')$ into circular buffer $\mathcal{M}$
  \EndFor
 
  \State Sample a random batch $\mathcal{B}$ from $\mathcal{M}$
  
  \ForAll {$(s_i, a_i, r_i, s_i') \in \mathcal{B}$}
    \State Compute target $q_i$ according to eq. \eqref{eq:loss_dqn}
    \State Clip $q_i$ with maximum of $0$
  \EndFor

  \State Perform one step of gradient descent on $\ell_\theta$ w.r.t. $\theta$:
  $$\ell_\theta(\mathcal{B}) = \sum_{i=1}^{|\mathcal{B}|} (q_i - Q^\theta(s_i, a_i))^2$$

  \State Update parameters $\phi := (1-\rho) \phi + \rho \theta$
\EndWhile
\end{algorithmic}
\end{algorithm}

\begin{algorithm}[t]
\caption{Environment simulation}
\label{alg:env}
\begin{algorithmic}
\State {Operator $\odot$ marks the element-wise multiplication.}
\Function{Step}{$e \in \mathcal{E}$, $a \in \mathcal{A}$}
    \If{$a \in \mathcal{A}_c$}
      \State $ r = 
        \begin{dcases*}
           0 &  if $a = e.y$ \\
          -1 &  if $a \neq e.y$
        \end{dcases*}$
      \State Reset $e$ with a new sample $(x, y, \emptyset)$ from a dataset
      \State Return $(\mathcal{T}, r)$
    \ElsIf{$a \in \mathcal{A}_f$}
      \State Add $a$ to set of selected features: $e.\mathcal{\bar{F}} = e.\mathcal{\bar{F}} \cup a$
      \State Create mask $m_i=1$ if $f_i \in \mathcal{\bar{F}}$ and $0$ otherwise
      \State Return $((e.x \odot m, m), -\lambda c(a))$
    \EndIf
\EndFunction
\end{algorithmic}
\end{algorithm}

\subsubsection{Other possible extensions}
Our method provides a flexible framework that can be extended in numerous ways. 
More than one external classifier can be included in our model. We hypothesize that the model would learn to adaptively query different classifiers that perform well for the current sample. 

Instead of the static reward of $-1$ for misclassification, one can provide actual misclassification costs for the particular domain and our method would find the optimal solution. One could even set different misclassification costs to different classes, leading to the domain of cost-sensitive classification.

\begin{figure*}[t]
  \centering
  \setlength{\tabcolsep}{0pt}
  \begin{tabular}{cccc}
    \includegraphics[width=0.25\linewidth]{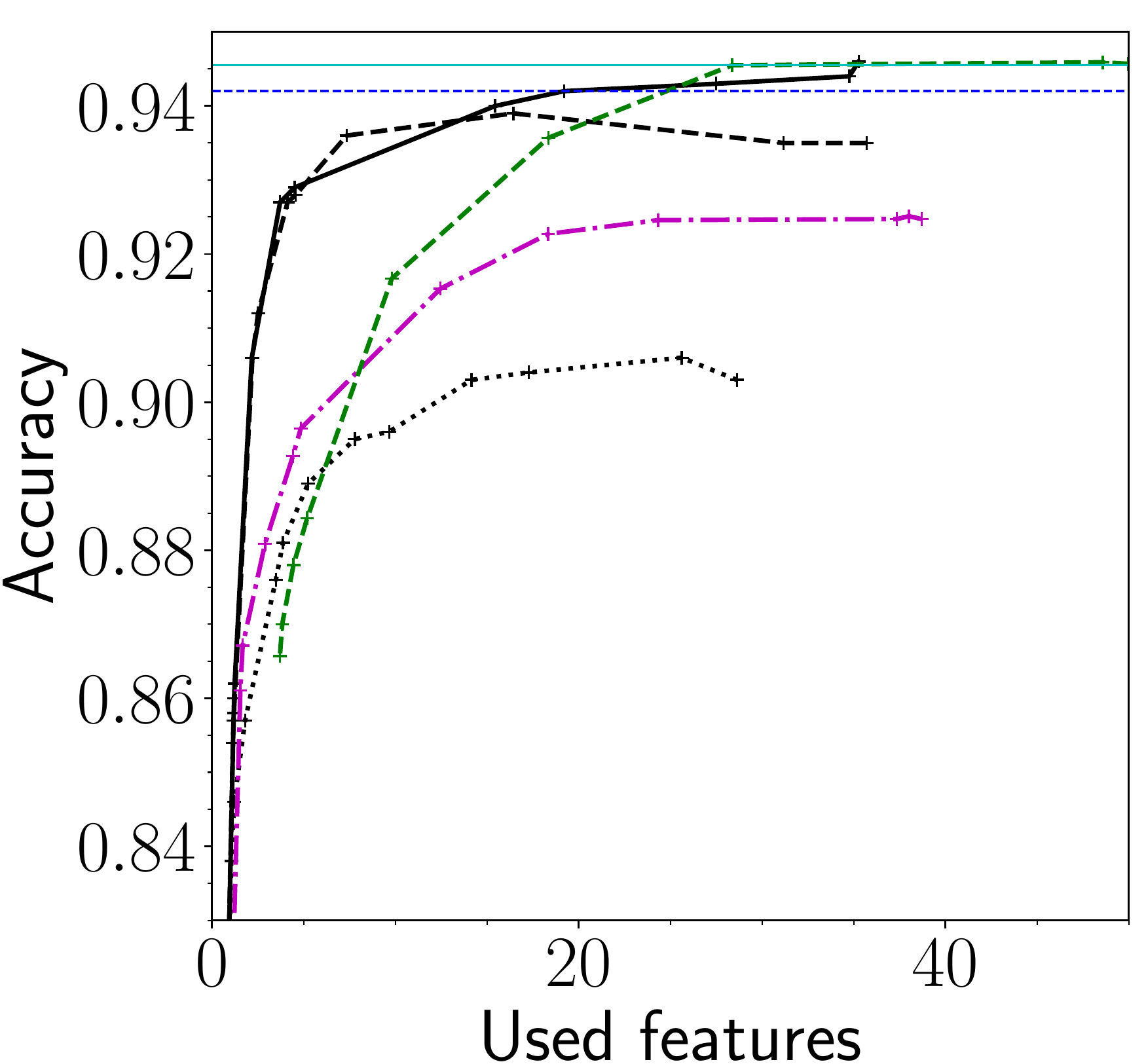} &
    \includegraphics[width=0.25\linewidth]{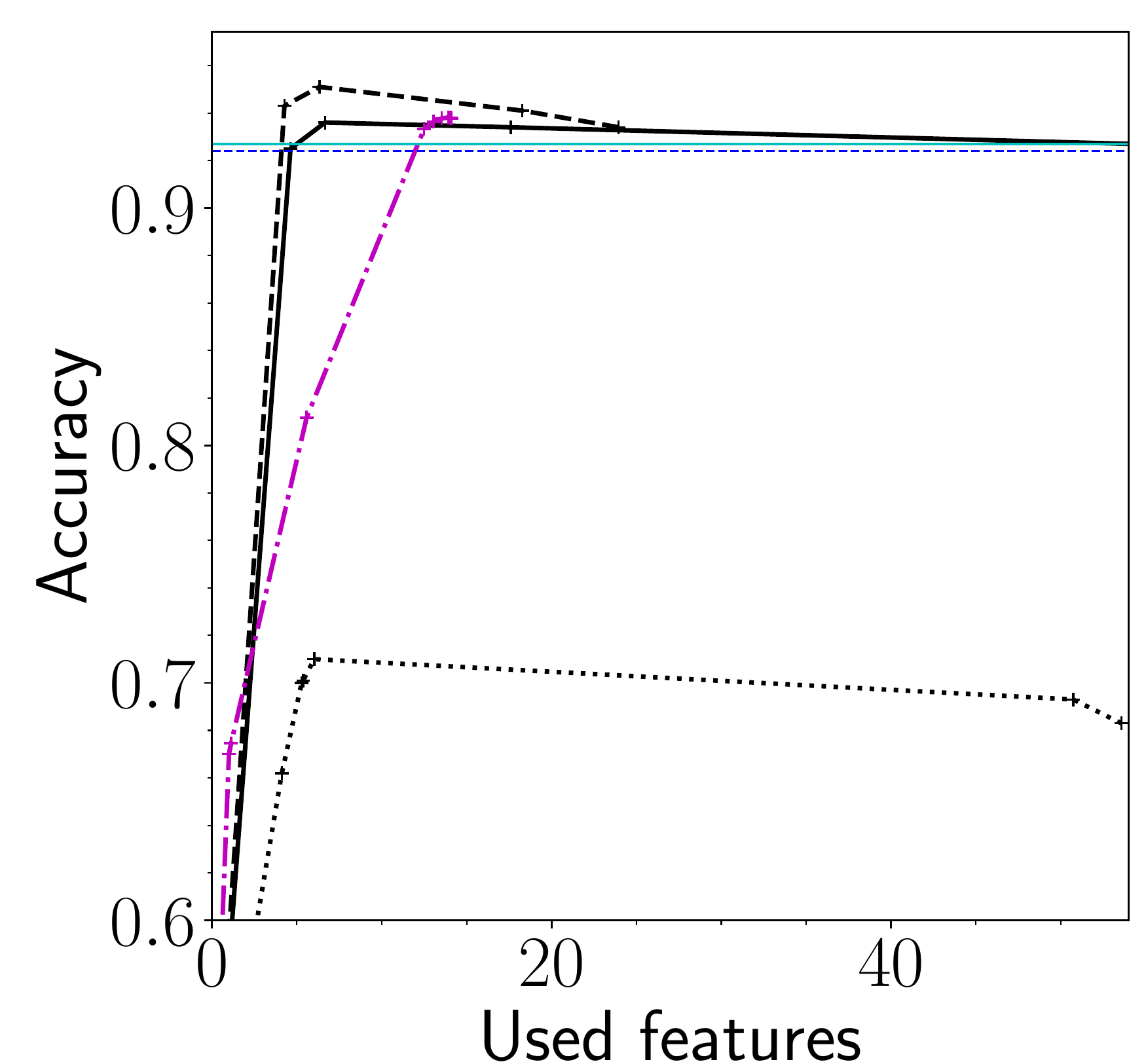} &
    \includegraphics[width=0.25\linewidth]{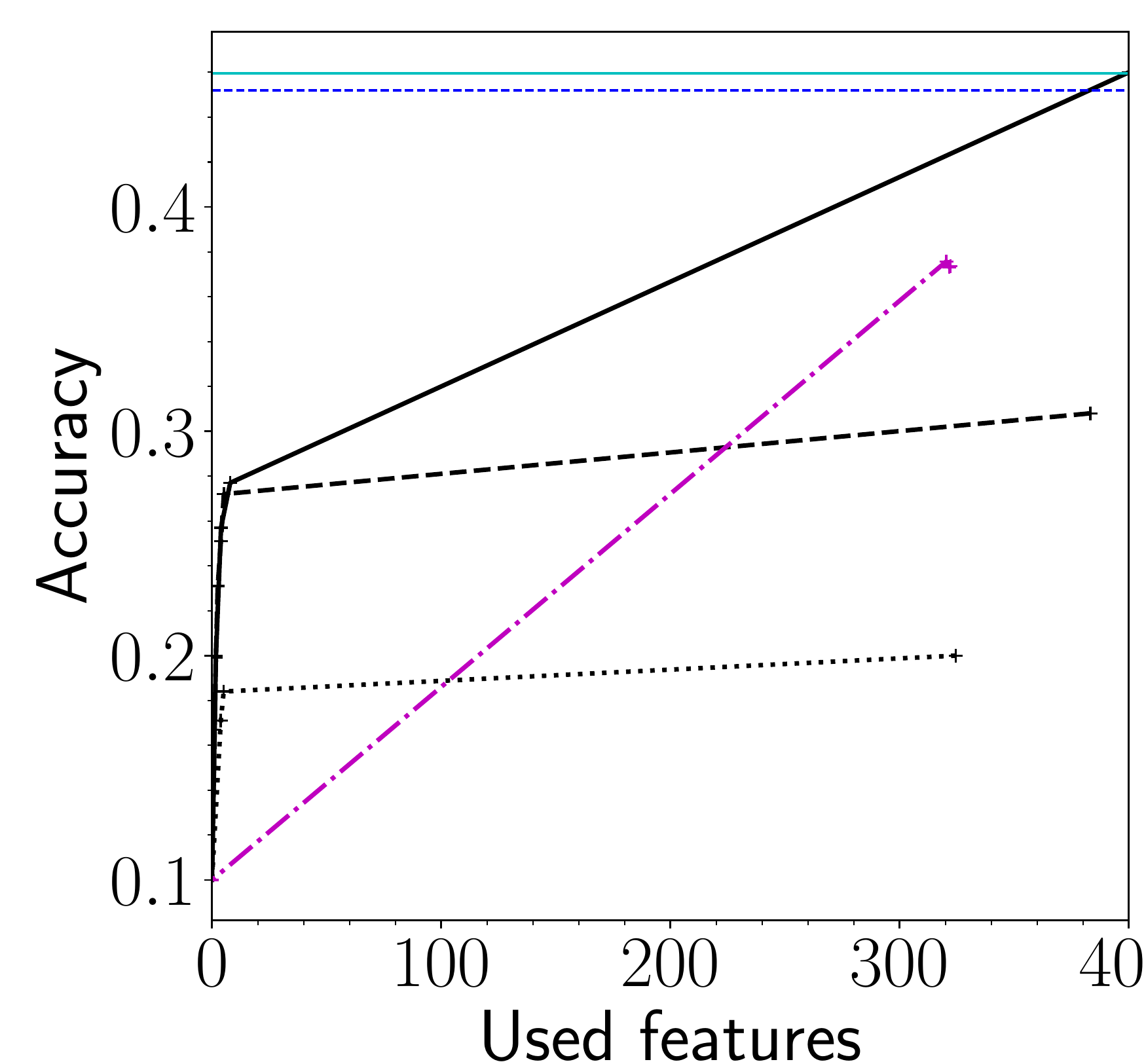} &
    \includegraphics[width=0.25\linewidth]{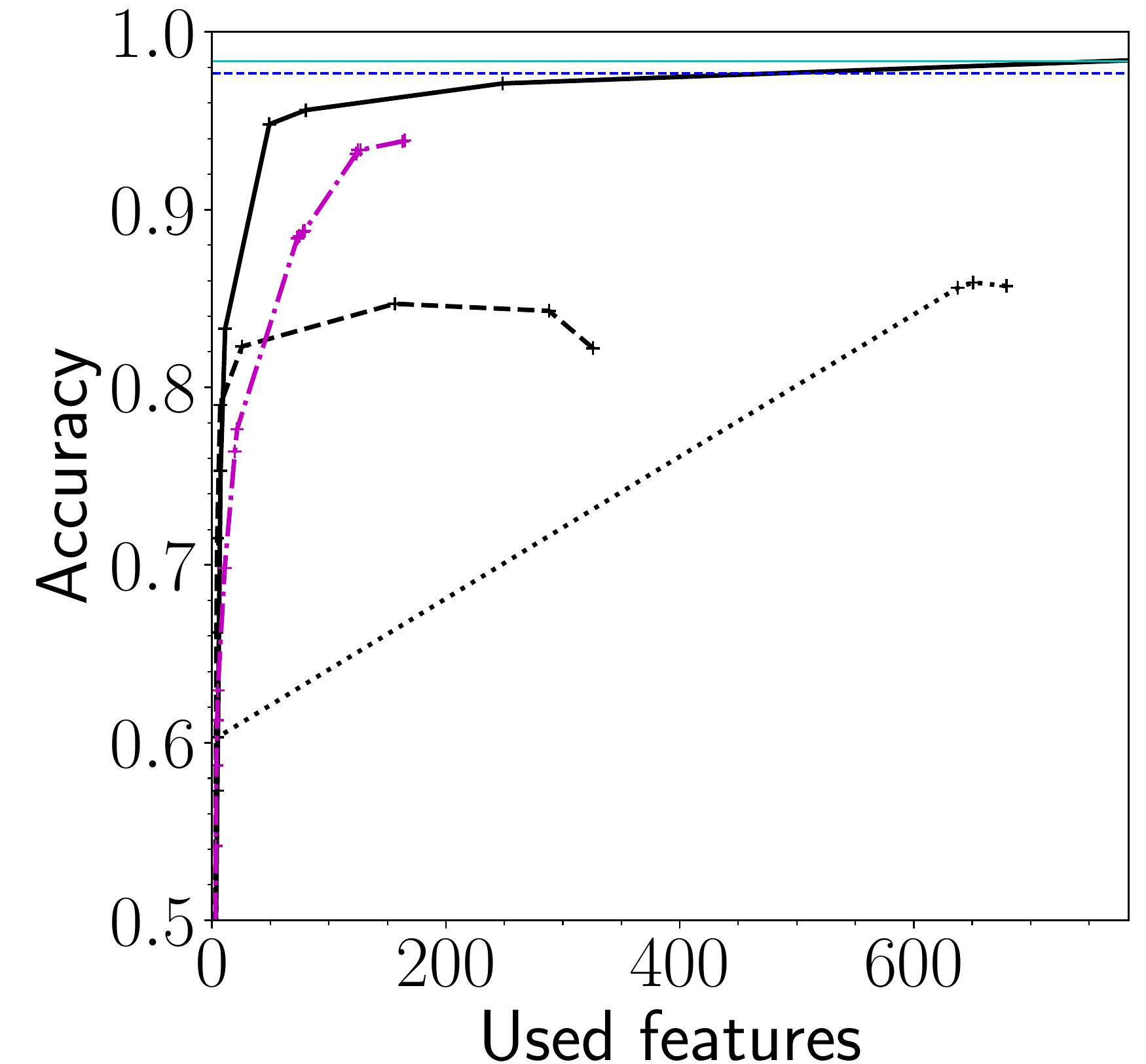} \\

    (a) miniboone & (b) forest & (c) cifar & (d) mnist \\

    \includegraphics[width=0.25\linewidth]{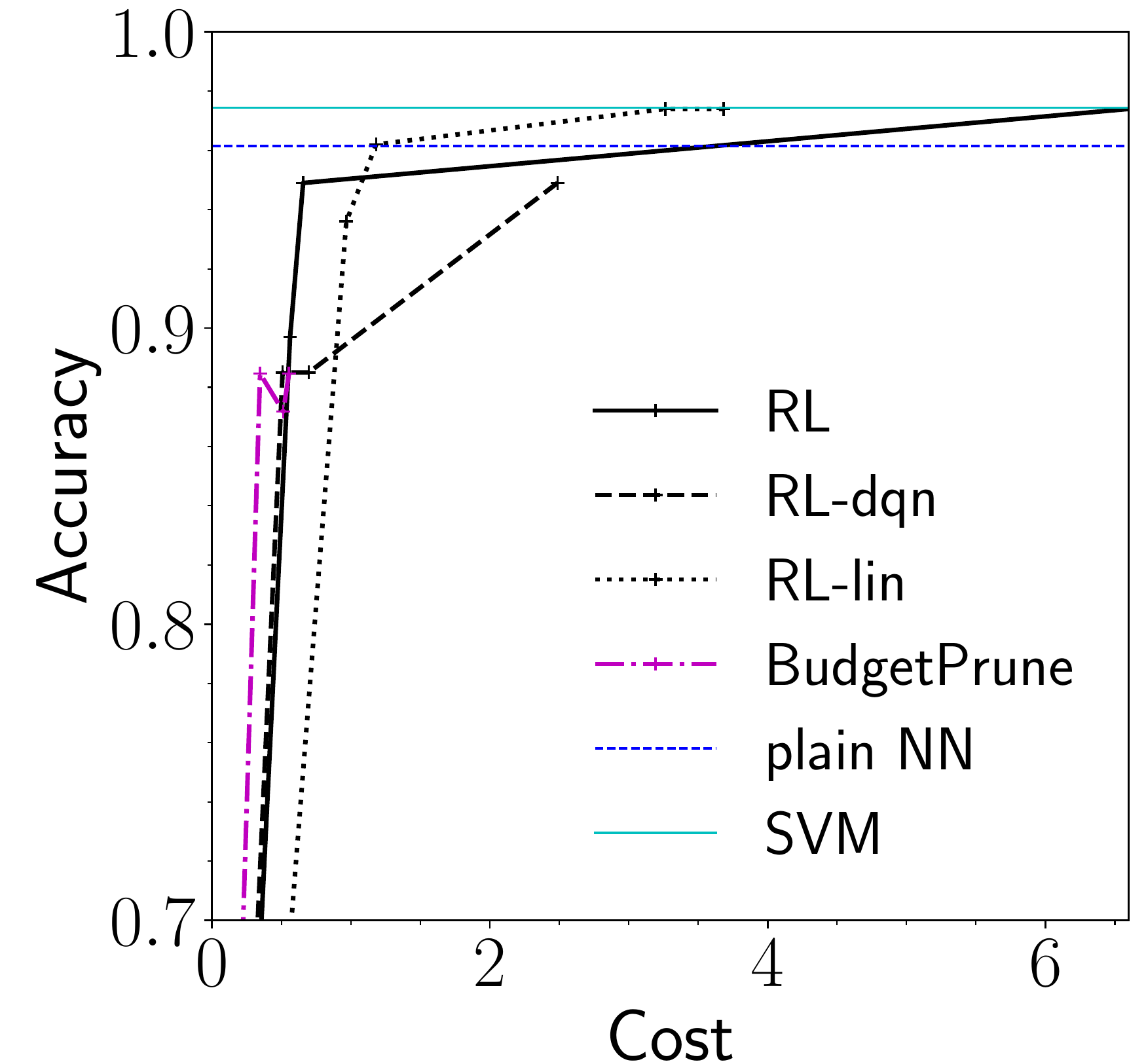} &
    \includegraphics[width=0.25\linewidth]{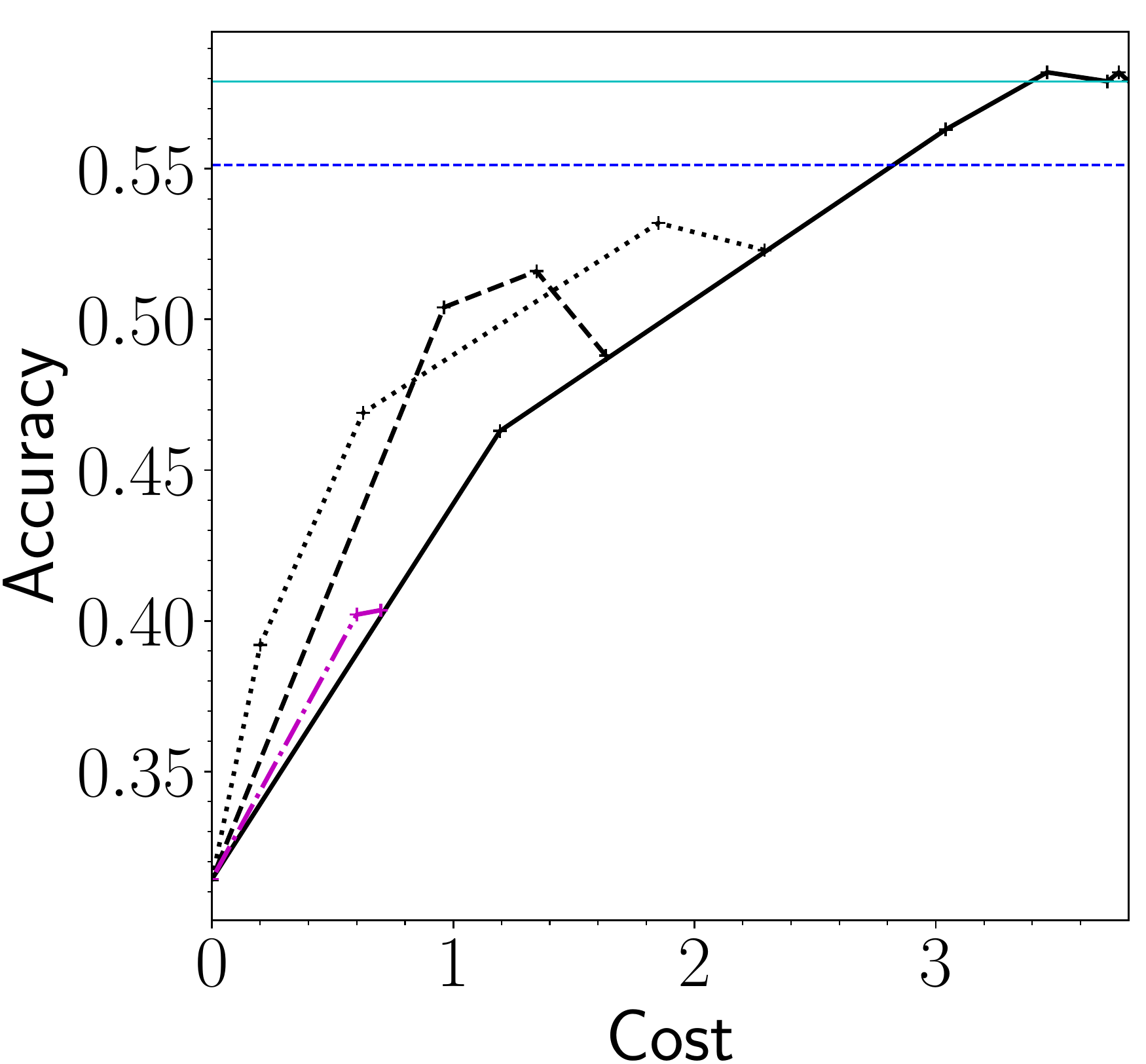} &
    \includegraphics[width=0.25\linewidth]{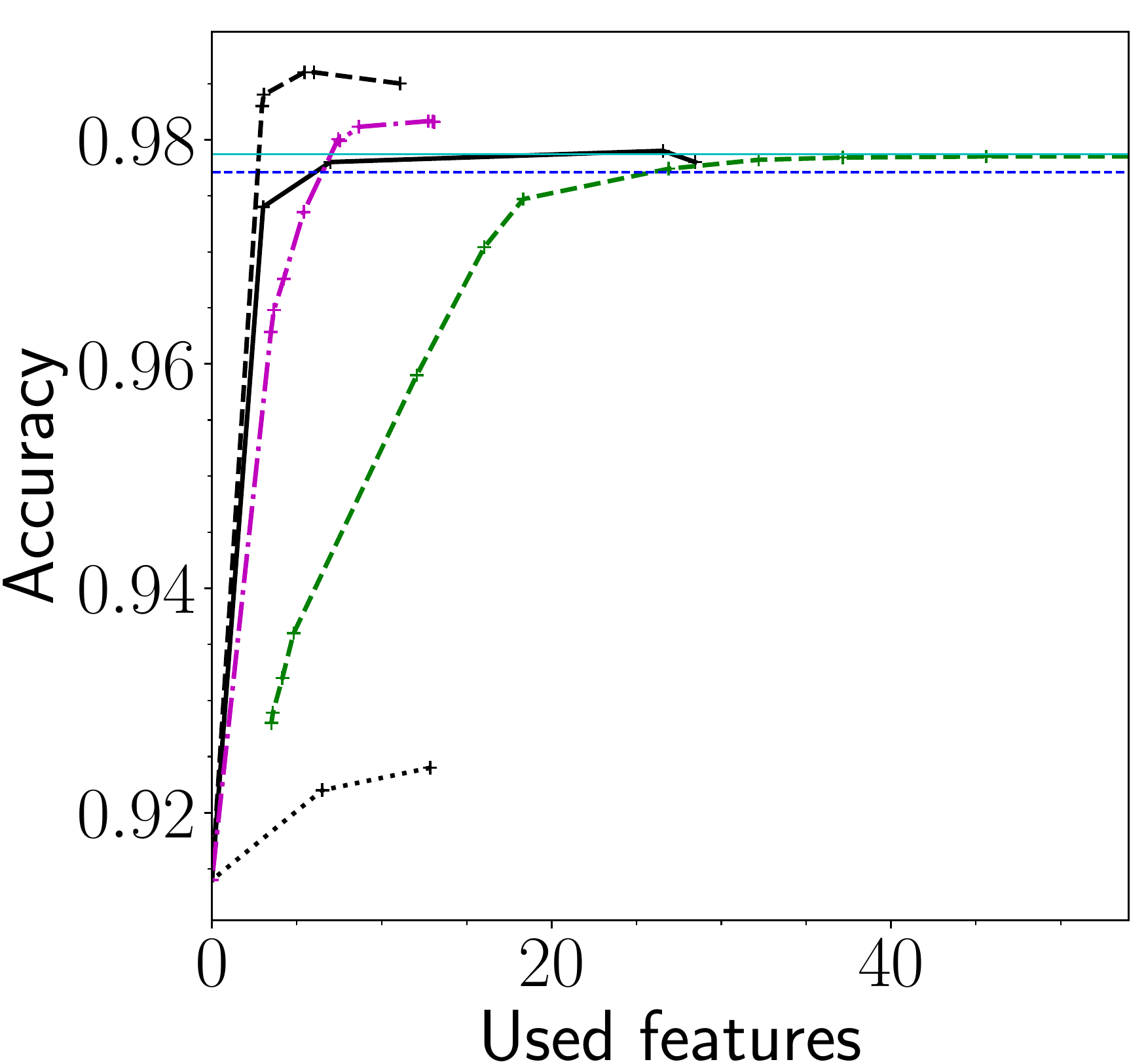} &
    \includegraphics[width=0.25\linewidth]{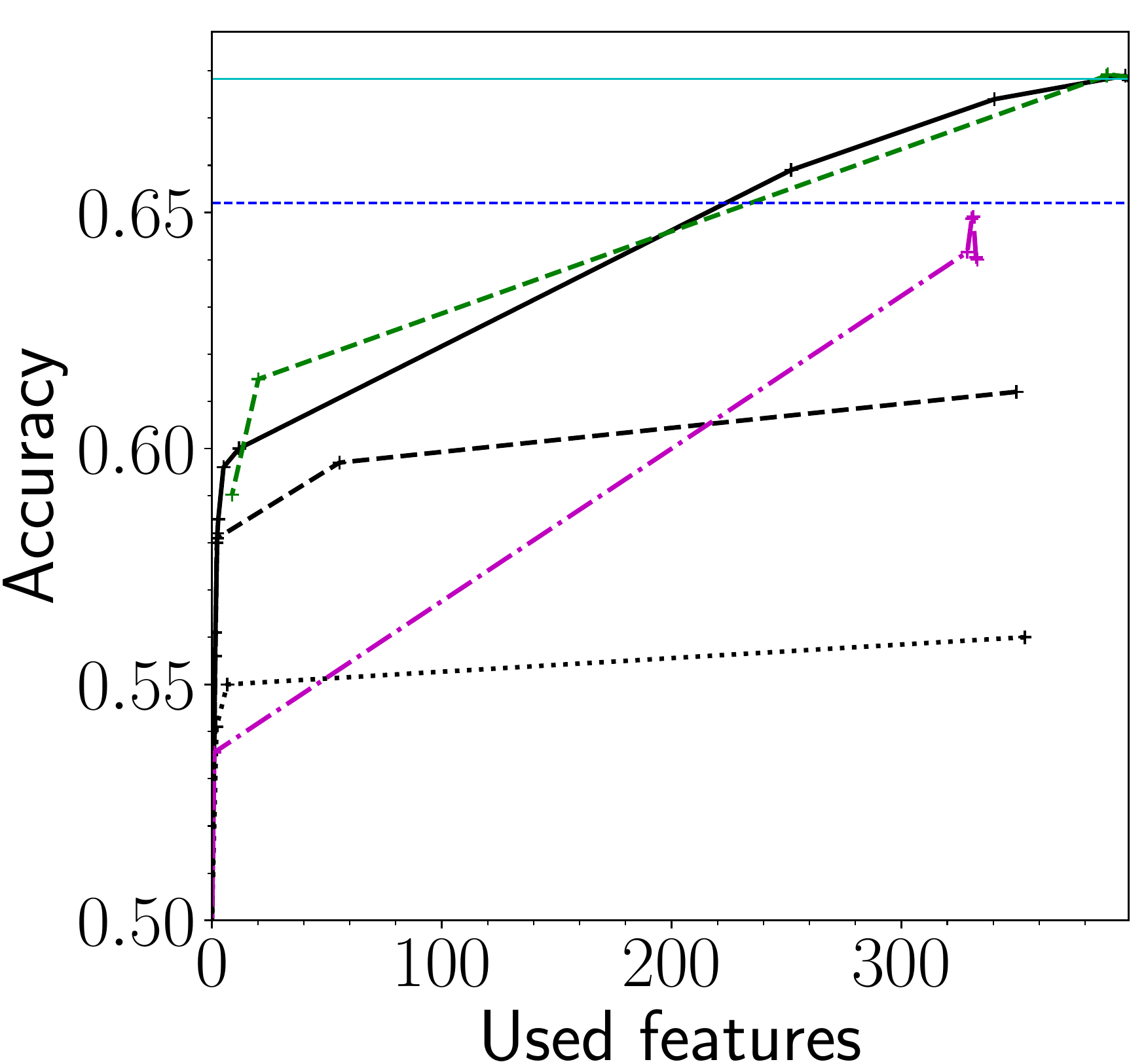} \\

    (e) wine & (f) yeast & (g) forest-2 & (h) cifar-2
  \end{tabular}

  \caption{Comparison with prior-art algorithms. Adapt-Gbrt is not shown on (b-f), because the open-sourced code is not compatible with multi-class datasets. RL-lin is Q-learning with linear approximation, RL-dqn is the baseline method. The horizontal lines correspond to the plain neural network based classifier and SVM. Datasets forest-2 and cifar-2 have two clases.}
  \label{fig:results}
\end{figure*}

In real-world cases, several features often come together with one request (e.g. one request to a cloud service can provide many values). In this case, the features can be grouped together to create a macro-feature, which is provided at once. In other cases, the number of the features may be too large to be practical. In these cases, existing algorithms \citep{wang2015efficient} can be used to automatically group the features.

\section{Experiments}

We compare our method with two state-of-the-art methods, \textit{Adapt-Gbrt} \citep{nan2017adaptive} and \textit{BudgetPrune} \citep{nan2016pruning} in Figure\;\ref{fig:results}. We instantiate our method in three versions: Q-learning with linear approximation (\textit{RL-lin}), Q-learning with neural networks (\textit{RL-dqn}) and the complete agent with all extensions (\textit{RL}).

\subsection{Experiment Details}
We use several public datasets \citep{uciml,krizhevsky2009learning}, which are summarized in Table\;\ref{tab:datasets}. We use three two-class and five multi-class datasets, with number of features ranging from 8 to 784. Two datasets were assigned random costs, selected from $\{0.2, 0.4, 0.6, 0.8\}$. Other datasets are left with the uniform cost of $1.0$ and the methods work as adaptive feature selection -- being able to select different features for different samples. We normalize the datasets with their mean and standard deviation and split them into training, validation and testing sets. Note that our results may differ from those reported in prior-art, because we use different preprocessing and splits.

\begin{table}[t]
    \begin{tabular}{lrrrrrc}
      \toprule
      Dataset     & feats.    & cls.      & \#trn     & \#val     &  \#tst    & costs     \\
      \midrule                                
      mnist       & 784       & 10        & 50k       & 10k       & 10k       & U         \\
      cifar       & 400       & 10        & 40k       & 10k       & 10k       & U         \\
      cifar-2     & 400       & 2         & 40k       & 10k       & 10k       & U         \\
      forest      & 54        & 7         & 200k      & 81k       & 300k      & U         \\
      forest-2    & 54        & 2         & 200k      & 81k       & 300k      & U         \\
      miniboone   & 50        & 2         & 45k       & 19k       & 65k       & U         \\
      wine        & 13        & 3         & 70        & 30        & 78        & V         \\
      yeast       & 8         & 10        & 600       & 200       & 684       & V         \\
      \bottomrule
    \end{tabular}

    \caption{Used datasets. The cost is either uniform (U) or variable (V) across features.}
    \label{tab:datasets}
\end{table}

Adapt-Gbrt is a random forest (RF) based algorithm and, similarly to our HPC extension, it uses an external HPC model. It jointly learns a gating function and Low-Prediction Cost model (LPC) that adaptively approximates HPC in regions where it makes accurate predictions. The gating function redirects the samples to either HPC or LPC. BudgetPrune is another algorithm that prunes an existing RF using linear programming to optimize the cost vs. accuracy trade-off. 

We obtained the results for both Adapt-Gbrt and BudgetPrune by running the source code published by their authors. The published version of Adapt-Gbrt is restricted to datasets with only two classes, hence we report it only on those. 

Separately, we train an SVM as the external classifier on each dataset. Hyperparameters were selected using cross-validation. This SVM is then used as the external classifier (HPC) both in our method and AdaptGbrt. We also train a plain neural network based classifier with the same size as our model for each dataset and report its performance, along with the performance of the SVM.

\begin{figure}[b]
    \centering
    \includegraphics[width=0.5\linewidth]{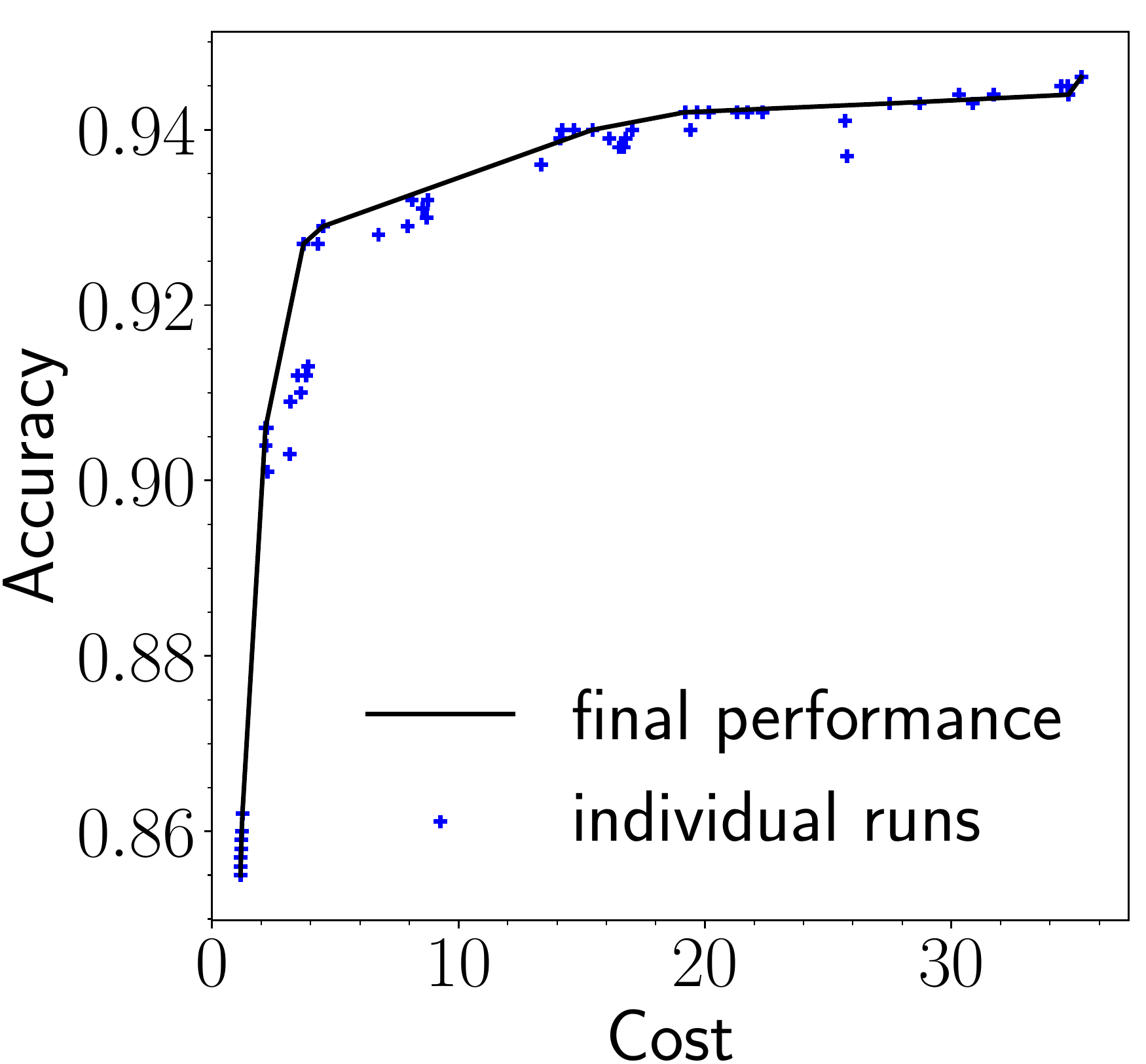}

    \caption{Different runs of our algorithm in the miniboone dataset. Validation set was used to select the best points.}
    \label{fig:mb_scatter}
\end{figure}

All algorithms include a $\lambda$-like parameter and we repeatedly run each of them with different settings and seeds. Following the methodology of \citet{nan2017adaptive}, we evaluate the runs on the validation set and select the points making a convex hull in the cost-accuracy space. Finally, we re-evaluate these points on the test set. The method corresponds to a best case scenario, but does not describe the variance between different runs. We show all runs in the miniboone dataset, along with the selected points in Figure\;\ref{fig:mb_scatter}.

We define a hyperparameter \textit{epoch length}, that differs across datasets, to be a number of training steps. For each dataset, we heuristically estimate the size of the neural network (NN) by training NN based classifiers with number of neurons selected from $\{64, 128, 256, 512\}$ in three layers with ReLu activation. We choose the lowest size that performs well on the task, without excess complexity. We use automatic learning-rate scheduling and early-stopping. We track the reward averaged over epochs on the training set and lower the learning-rate immediately when the reward fails to increase. Similarly, we use the average reward on the validation set and interrupt the computation if the reward fails to increase consecutively for three epochs. \textbf{We keep other hyperparameters the same across all experiments}, and their exact values are reported in Tables \ref{tab:global-parameters}, \ref{tab:dataset-parameters} in Appendix. 

\begin{figure}[t]
  \centering
  \setlength{\tabcolsep}{0pt}
  \begin{tabular}{cc}
    \includegraphics[width=0.5\linewidth]{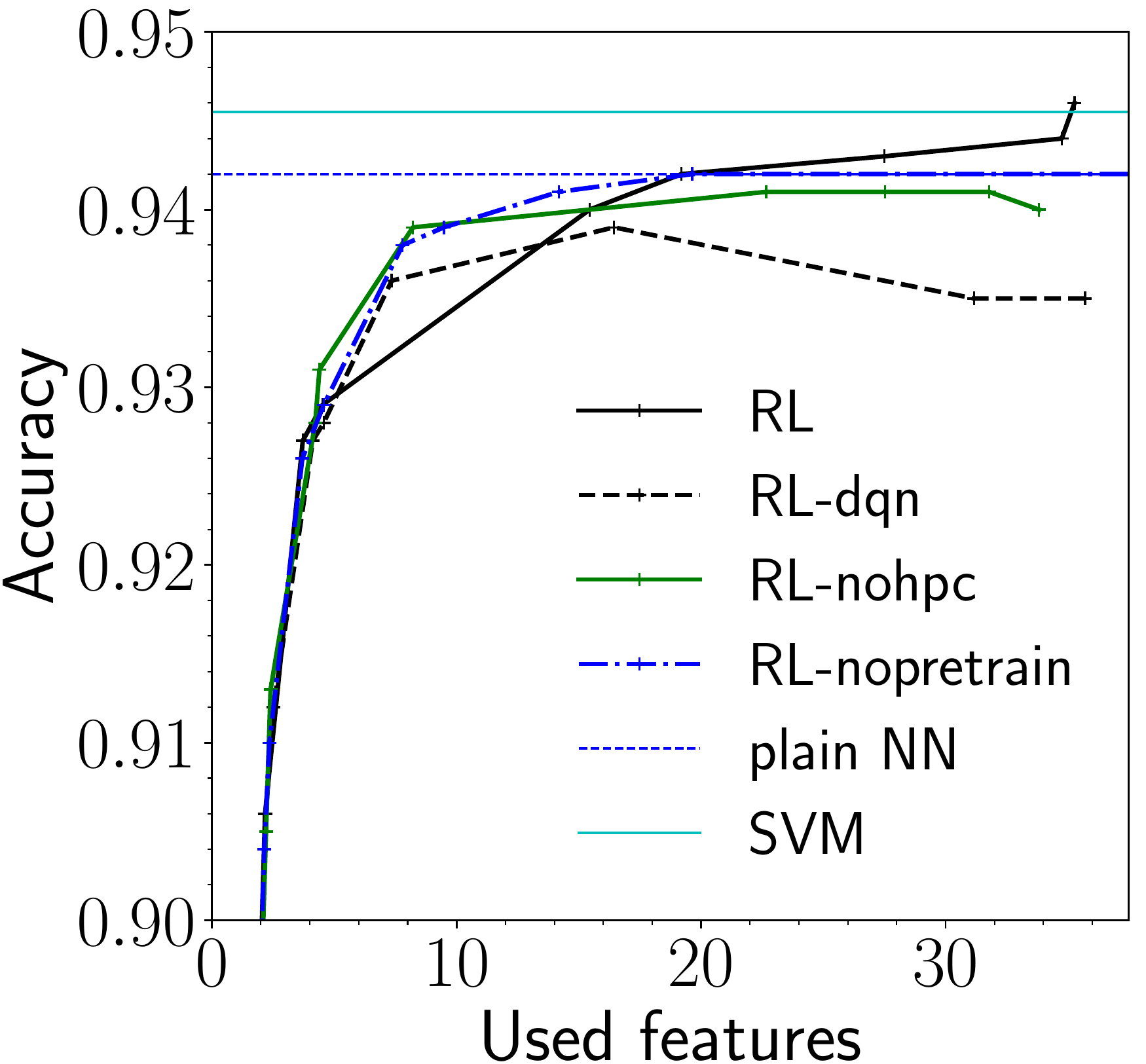}  &
    \includegraphics[width=0.5\linewidth]{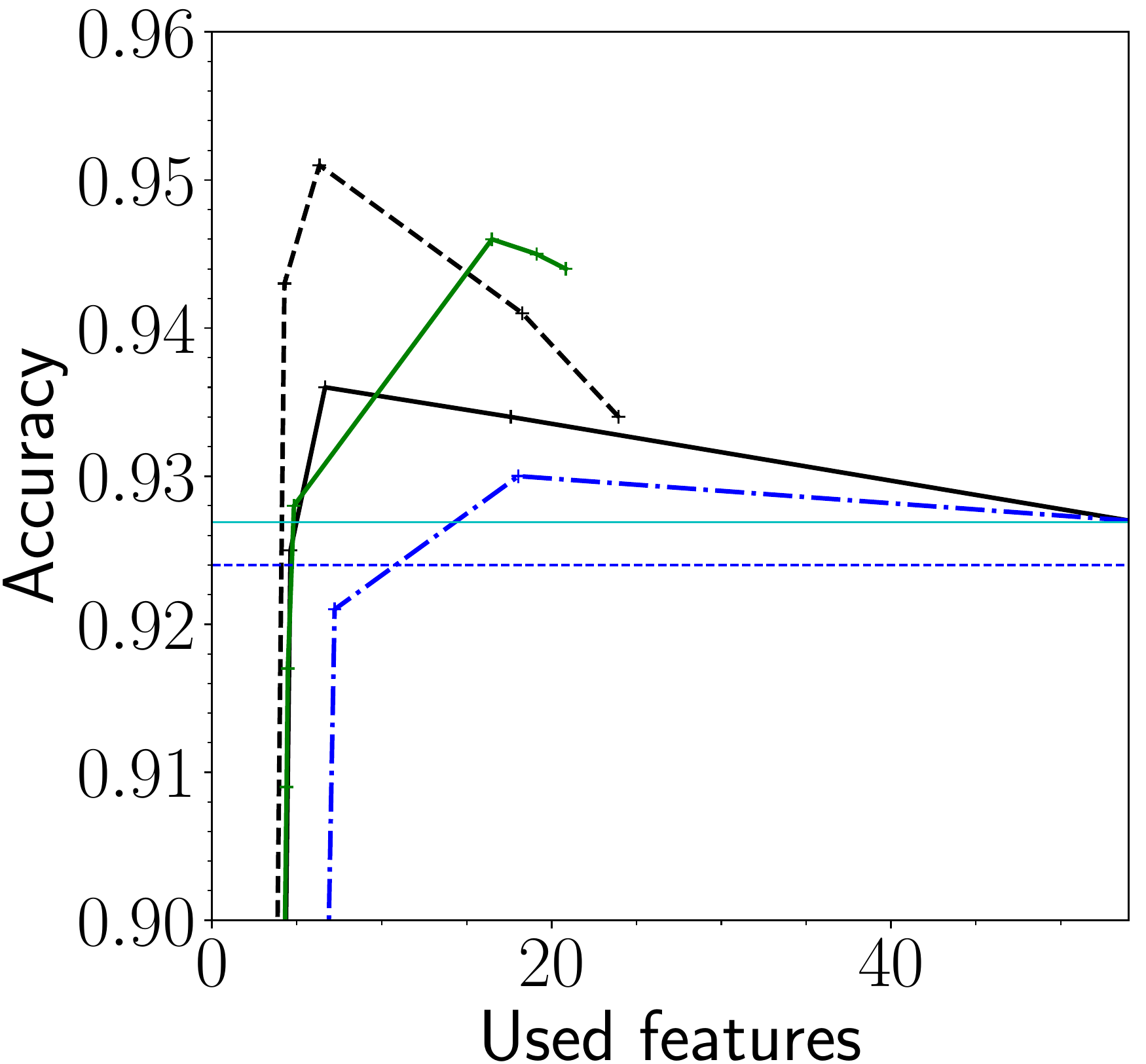} \\
    (a) miniboone & (b) forest
  \end{tabular}

  \caption{Performance of different versions of the algorithm.}
  \label{fig:versions}
\end{figure}

\subsection{Discussion}
Comparing the full agent (RL) to the prior-art, the RL outperforms Adapt-Gbrt and BudgetPrune in miniboone, forest (except for the very low cost range), cifar, mnist and cifar. On cifar-2 and forest-2, there are ranges of costs, where another algorithm is better, but it is a different algorithm each time. Minding the fact that the algorithm was not specifically tuned for each dataset, this indicates high robustness.

A notable result is that the step from linear approximation (RL-lin) to neural networks (RL-dqn) translates into a large improvement of performance in all tested datasets but yeast and wine. These two datasets are small and the linear method works well. Because the linear approximation is a special case of neural networks, it indicates that the results can be improved if the architecture and hyperparameters were adjusted separately for each dataset. The additional extensions of the full agent (RL) contributed mainly to the stability of the agent, which was rendered into better performance in the large datasets: cifar, cifar-2 and mnist (Figure\;\ref{fig:results}cdh).

\begin{figure}[t]
  \centering
  \includegraphics[width=1.0\linewidth]{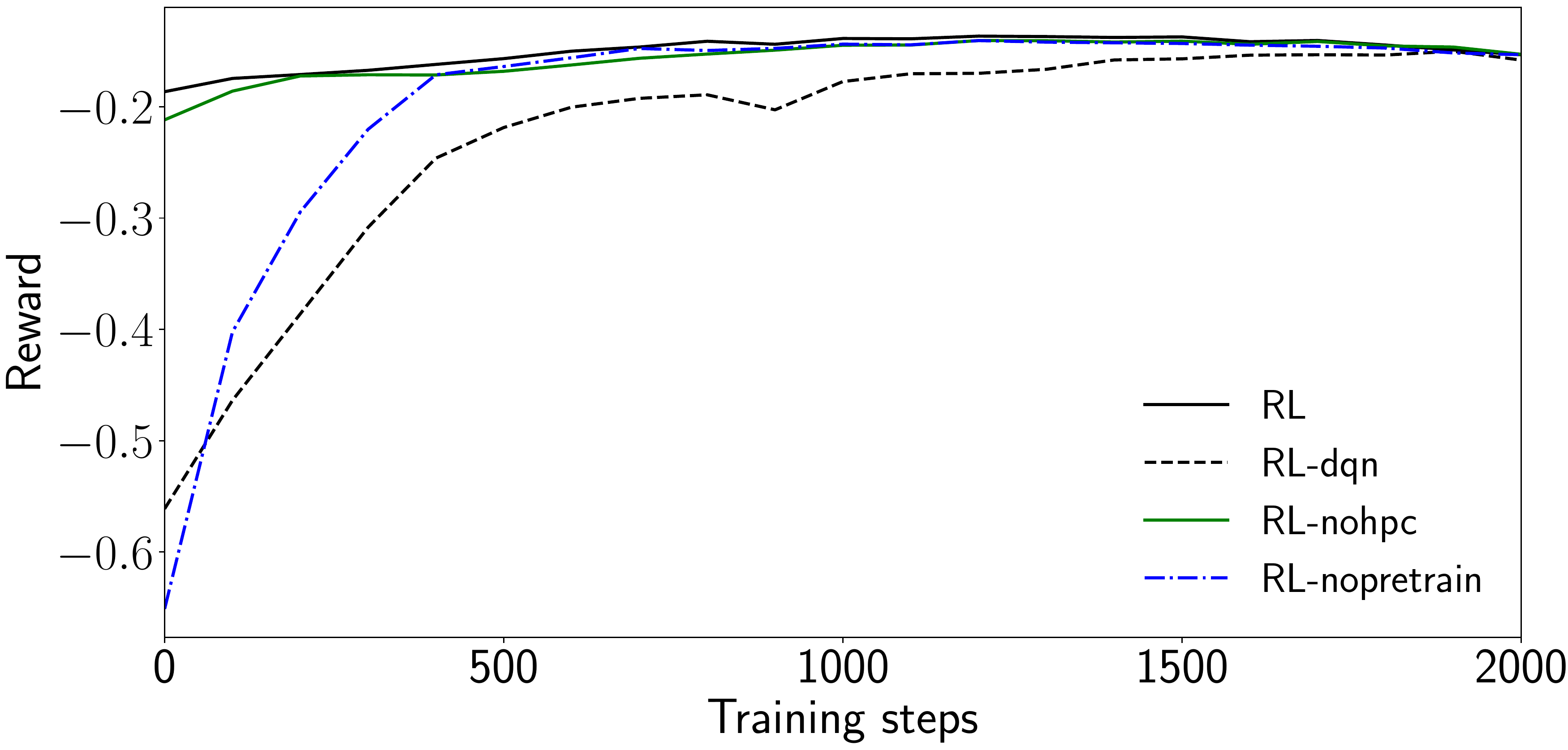}

  \caption{Average reward during training. All versions converge to a good solution, but with different speed. Miniboone, $\lambda=0.01$, averaged over 10 runs.}
  \label{fig:progress}
\end{figure}

We further investigate the different versions of our algorithm in Figure\;\ref{fig:versions} and \ref{fig:progress}. We found that the there are differences in convergence speeds, which is another important factor beside performance. Without advanced RL techniques, RL-dqn converges slowest. With pretraining, the model is initialized closer to an optima. Another fact is that the performance improves continuously with progression of the training. This may be useful in non-stationary domains, where the statistics of the domain change in time. Also, the learning can be interrupted anytime, still providing a sensible result. 

\begin{figure}[t]
  \centering
  \setlength{\tabcolsep}{0pt}
  \begin{tabular}{cc}
    \includegraphics[width=0.5\linewidth]{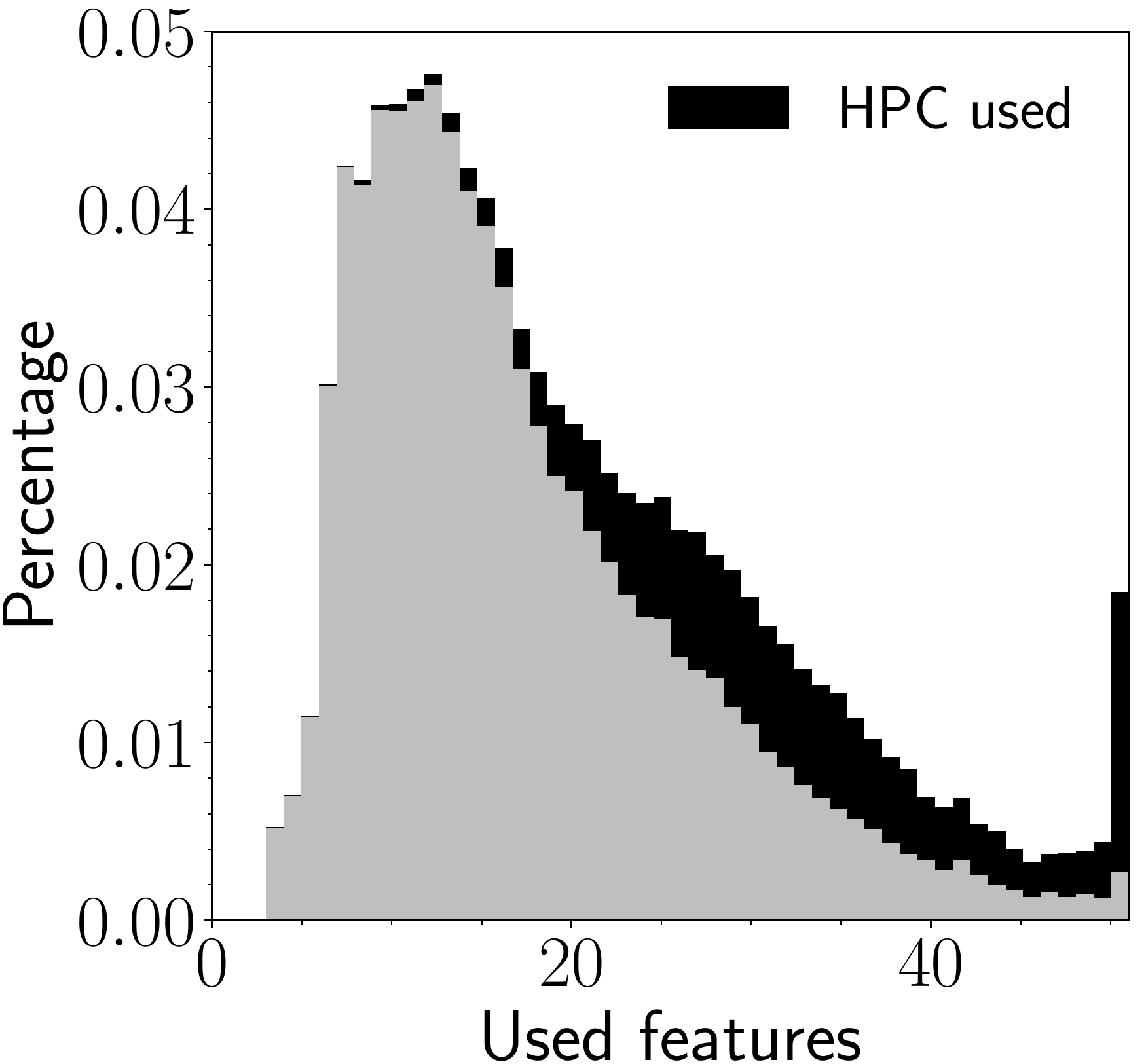}  &
    \includegraphics[width=0.5\linewidth]{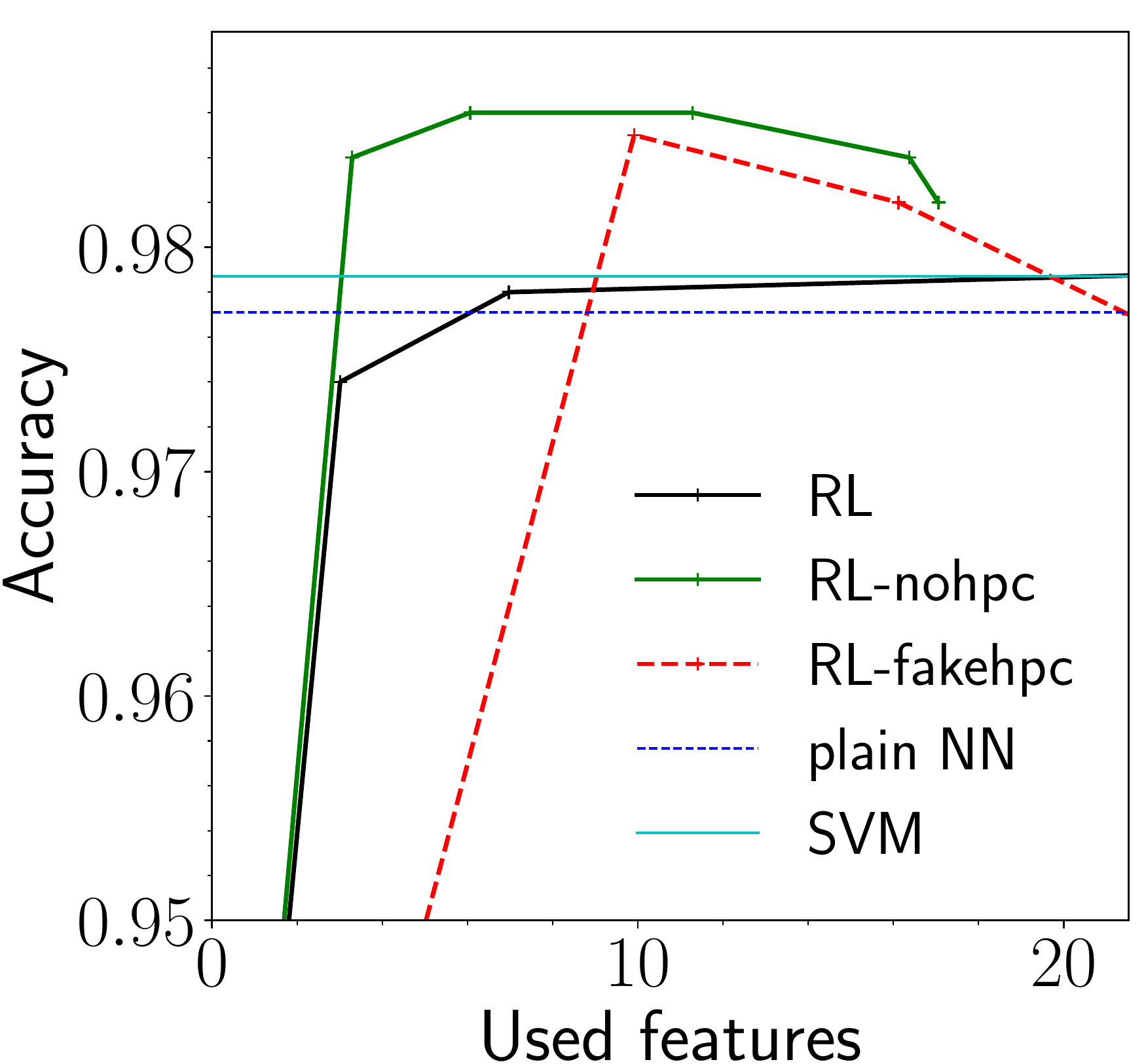} \\

    (a) Number of used features & (b) HPC analysis; forest-2
  \end{tabular}

  \caption{(a) Histogram of number of used features across the whole miniboone dataset, black indicates that the HPC was queried at that point. (b) Comparison of the full agent, an agent without HPC and a non-predictive classifier on the forest-2 dataset. The used SVM has very low training error, causing the RL agent to overfit. The agent is able to ignore the fake HPC to some extent.}
  \label{fig:analysis}
\end{figure}

Another interesting fact is that RL-dqn and the version without HPC perform better on the forest dataset. The situation is even more profound in the case of forest-2 (Figure\;\ref{fig:results}g), where the BudgetPrune also outperforms the SVM. We investigated the issue and found that the SVM has 0.995 accuracy on the training set of forest-2, causing the RL agent to overfit and rely on its HPC. To further analyze the issue, we trained models without the external classifier (\textit{RL-nohpc}) and with a non-predictive HPC (\textit{RL-fakehpc}), by preparing random predictions for all samples (see Figure\;\ref{fig:analysis}b). We found that the agent without the external classifier performed better than the agent with HPC, confirming that the agent indeed overfits. With the case of the non-predictive HPC, it introduced some instability into the training, but the performance is still higher for a range of costs. The results show that including HPC should be done with caution -- the agent is able to ignore a non-predictive classifier to some extent, but an overfit classifier can be harmful. Surprisingly, the agent without HPC module is not constrained by the performance of the plain neural network, indicating a good generalization ability.

In Figure\;\ref{fig:analysis}a we summarize how many features a trained agent requested for different samples in the miniboone dataset. We include the HPC queries to get intuition about in which cases the external classifier is used. The agent classifies 40\% of all samples with under 15 features and with almost no HPC usage. For the rest of samples, the agent queries the HPC in about 19\% cases. The histogram confirms that the agent is classifies adaptively, requesting a different amount of features across samples, and triggering the external classifier on demand. With further analysis, we confirm that the agent also requests a \textit{different} set of features for different samples.

\section{Related Work}
We build upon work of \citet{dulac2011datum}, which used Q-learning with linear regression, resulting in limited performance. We replace the linear approximation with neural networks, extend the approach with supervised pretraining and external HPC and thoroughly compare to recent methods.

\citet{contardo2016recurrent} use a recurrent neural network that uses attention to select blocks of features and classifies after a fixed number of steps. Compared to our work, decisions of this network are hard to explain. On the other hand, our Q-learning based algorithm produces semantically meaningful values for each action. Moreover, as we use a standard algorithm, we can directly benefit from any enhancements made by the community.

There is a plethora of tree-based algorithms \citep{xu2012greedy,kusner2014feature,xu2013cost,xu2014classifier,nan2015feature,nan2016pruning,nan2017adaptive}. Last two articles implement BudgetPrune and AdaptGbrt algorithms, that we describe in Experiments.

A different set of algorithms employed linear programming (LP) to this domain \citep{wang2014lp,wang2014model}. \citet{wang2014model} use LP to select a model with the best accuracy and lowest cost, from a set of pre-trained models, all of which use a different set of features. The algorithm also chooses a model based on the complexity of the sample, similarly to Adapt-Gbrt.

\citet{wang2015efficient} propose to reduce the problem by finding different disjoint subsets of features, that are used together as macro-features. These macro-features form a graph, which is solved with dynamic programming. The algorithm for finding different subsets of features is complementary to our algorithm and could be possibly used jointly to improve performance.

\citet{trapeznikov2013supervised} use a fixed order of features to reveal, with increasingly complex models that can use them. However, the order of features is not computed, and it is assumed that it is set manually. Our algorithm is not restricted to a fixed order of features (for each sample it can choose a completely different subset), and it can also find their significance automatically.

Recent work \citep{maliah2018mdp} focuses on CwCF with misclassification costs, constructs decision trees over attribute subsets and use their leaves to form states of an MDP. They directly solve the MDP with value-iteration for small datasets with number of features ranging from 4-17. On the other hand, our method can be used to find an approximate solution to much larger datasets. In this work, we do not account for misclassification costs, but they could be easily incorporated into the rewards for classification actions.

\citet{tan1993cost} analyzes a problem similar to our definition, but algorithms introduced there require memorization of all training examples, which is not scalable in many domains.

\section{Conclusion}
Classification with costly features (CwCF) is a sequential decision-making problem that inspired many problem-specific solutions. While domain independent methods for general decision-making problems, such as reinforcement learning (RL), have been proposed as a solution, they have not performed well in comparison to the state-of-the-art algorithms specifically tailored to this problem. In light of the recent progress, we revisit the RL approach and evaluate several Deep Q-learning based algorithms. The investigation shows that even basic Deep RL is comparable and often superior to state-of-the-art algorithms Adapt-Gbrt \citep{nan2017adaptive} and BudgetPrune \citep{nan2016pruning}. 

The RL approach has a relatively low number of parameters and is robust with respect to their modification, which facilitates the deployment. It is a modular framework, allowing for numerous extensions, for example the supervised pre-training and the inclusion of an external classifier demonstrated in this paper. RL approach can also benefit from future advancements of Deep RL itself, which we demonstrated by utilizing Double Q-learning, Dueling Architecture and Retrace. Lastly, using RL for CwCF enables continual adaptation to changes in non-stationary environments, commonly present in real-world problems.

\section*{Acknowledgments}
This research was supported by the European Office of Aerospace Research and Development (grant no. FA9550-18-1-7008) and by The Czech Science Foundation (grants no. 18-21409S and 18-27483Y). The GPU used for this research was donated by the NVIDIA Corporation. Computational resources were provided by the CESNET LM2015042 and the CERIT Scientific Cloud LM2015085, provided under the program Projects of Large Research, Development, and Innovations Infrastructures.

\section*{Appendix}
\renewcommand{\thetable}{A.\arabic{table}}
\setcounter{table}{0}

\begin{table}[h]
    \begin{tabular}{llr}
      \toprule
      Symbol                          & description                                               & value               \\
      \midrule
      $|\mathcal{E}|$                 & no. of parallel environments                           & 1000                 \\                                                                    
      $\gamma$                        & discount-factor                                           & 1.0                 \\
      Retrace-$\lambda$               & Retrace parameter $\lambda$                               & 1.0                 \\
      $\rho$                          & target network update factor                              & 0.1                 \\
      $|\mathcal B|$                  & no. of steps in batch                                    & 50k                 \\
      $|\mathcal M|$                  & no. of episodes in memory                                & 40k                 \\
      $\epsilon_\text{-start}$        & starting exploration                                    & 1.0                 \\
      $\epsilon_\text{-end}$          & final exploration                                       & 0.1                 \\
      $\eta_\text{-start}$            & start greediness of policy $\pi$   & 0.5                 \\
      $\eta_\text{-end}$              & final greediness of policy $\pi$    & 0.0                 \\
      $\epsilon_\text{steps}$         & length of exploration phase                               & $2 \times \text{ep\_len}$ \\
      LR-pretrain                     & pre-training learning-rate                                & $1 \times 10^{-3}$\\
      LR-start                        & initial learning-rate                                     & $5 \times 10^{-4}$\\
      LR-min                          & minimal learning-rate                                     & $1 \times 10^{-7}$\\
      LR-scale                        & learning-rate multiplicator                               & 0.3                 \\
      \bottomrule
    \end{tabular}

    \caption{Global parameters}
    \label{tab:global-parameters}
\end{table}

\begin{table}[h]
    \centering
    \begin{tabular}{lrr}
      \toprule
      Dataset                 & hidden layer size & epoch length (\textit{ep\_len}) \\
      \midrule                                
      mnist\textsuperscript{\dag}                   & 512       & 10k      \\
      cifar\textsuperscript{\dag}                   & 512       & 10k      \\
      cifar-2\textsuperscript{\dag}                 & 512       & 10k      \\
      forest                  & 256       & 10k      \\
      forest-2                & 256       & 10k      \\
      miniboone               & 128       & 1k       \\
      wine                    & 128       & 1k       \\
      yeast                   & 128       & 1k       \\
      \bottomrule
    \end{tabular}

    \caption{Dataset parameters. \textsuperscript{\dag}Specific settings are used: $|\mathcal M|=10\text{k}, \text{LR-pretrain}=2 \times 10^{-5}, \text{LR-start}=10^{-5}$}
    \label{tab:dataset-parameters}
\end{table}

\small
\bibliographystyle{abbrvnat}
\bibliography{citations}

\begin{thebibliography}{24}
\providecommand{\natexlab}[1]{#1}
\providecommand{\url}[1]{\texttt{#1}}
\expandafter\ifx\csname urlstyle\endcsname\relax
  \providecommand{\doi}[1]{doi: #1}\else
  \providecommand{\doi}{doi: \begingroup \urlstyle{rm}\Url}\fi

\bibitem[Contardo et~al.(2016)Contardo, Denoyer, and
  Artieres]{contardo2016recurrent}
G.~Contardo, L.~Denoyer, and T.~Artieres.
\newblock Recurrent neural networks for adaptive feature acquisition.
\newblock In \emph{International Conference on Neural Information Processing},
  pages 591--599. Springer, 2016.

\bibitem[Dulac-Arnold et~al.(2011)Dulac-Arnold, Denoyer, Preux, and
  Gallinari]{dulac2011datum}
G.~Dulac-Arnold, L.~Denoyer, P.~Preux, and P.~Gallinari.
\newblock Datum-wise classification: a sequential approach to sparsity.
\newblock In \emph{Joint European Conference on Machine Learning and Knowledge
  Discovery in Databases}, pages 375--390. Springer, 2011.

\bibitem[Kingma and Ba(2015)]{kingma2014adam}
D.~P. Kingma and J.~Ba.
\newblock Adam: A method for stochastic optimization.
\newblock In \emph{International Conference on Learning Representations}, 2015.

\bibitem[Krizhevsky and Hinton(2009)]{krizhevsky2009learning}
A.~Krizhevsky and G.~Hinton.
\newblock Learning multiple layers of features from tiny images.
\newblock Master's thesis, University of Toronto, 2009.

\bibitem[Kusner et~al.(2014)Kusner, Chen, Zhou, Xu, Weinberger, and
  Chen]{kusner2014feature}
M.~Kusner, W.~Chen, Q.~Zhou, Z.~Xu, K.~Weinberger, and Y.~Chen.
\newblock Feature-cost sensitive learning with submodular trees of classifiers.
\newblock In \emph{AAAI Conference on Artificial Intelligence}, pages
  1939--1945, 2014.

\bibitem[Lichman(2013)]{uciml}
M.~Lichman.
\newblock {UCI} machine learning repository, 2013.
\newblock URL \url{http://archive.ics.uci.edu/ml}.

\bibitem[Lillicrap et~al.(2016)Lillicrap, Hunt, Pritzel, Heess, Erez, Tassa,
  Silver, and Wierstra]{lillicrap2015continuous}
T.~P. Lillicrap, J.~J. Hunt, A.~Pritzel, N.~Heess, T.~Erez, Y.~Tassa,
  D.~Silver, and D.~Wierstra.
\newblock Continuous control with deep reinforcement learning.
\newblock In \emph{International Conference on Learning Representations}, 2016.

\bibitem[Lin(1993)]{lin1993reinforcement}
L.-J. Lin.
\newblock \emph{Reinforcement learning for robots using neural networks}.
\newblock PhD thesis, Carnegie Mellon University, 1993.

\bibitem[Maliah and Shani(2018)]{maliah2018mdp}
S.~Maliah and G.~Shani.
\newblock Mdp-based cost sensitive classification using decision trees.
\newblock In \emph{AAAI Conference on Artificial Intelligence}, pages
  3746--3753, 2018.

\bibitem[Mnih et~al.(2015)Mnih, Kavukcuoglu, Silver, Rusu, Veness, Bellemare,
  Graves, Riedmiller, Fidjeland, Ostrovski, et~al.]{mnih2015human}
V.~Mnih, K.~Kavukcuoglu, D.~Silver, A.~A. Rusu, J.~Veness, M.~G. Bellemare,
  A.~Graves, M.~Riedmiller, A.~K. Fidjeland, G.~Ostrovski, et~al.
\newblock Human-level control through deep reinforcement learning.
\newblock \emph{Nature}, 518\penalty0 (7540):\penalty0 529--533, 2015.

\bibitem[Munos et~al.(2016)Munos, Stepleton, Harutyunyan, and
  Bellemare]{munos2016safe}
R.~Munos, T.~Stepleton, A.~Harutyunyan, and M.~Bellemare.
\newblock Safe and efficient off-policy reinforcement learning.
\newblock In \emph{Advances in Neural Information Processing Systems}, pages
  1054--1062, 2016.

\bibitem[Nan and Saligrama(2017)]{nan2017adaptive}
F.~Nan and V.~Saligrama.
\newblock Adaptive classification for prediction under a budget.
\newblock In \emph{Advances in Neural Information Processing Systems}, pages
  4730--4740, 2017.

\bibitem[Nan et~al.(2015)Nan, Wang, and Saligrama]{nan2015feature}
F.~Nan, J.~Wang, and V.~Saligrama.
\newblock Feature-budgeted random forest.
\newblock In \emph{International Conference on Machine Learning}, pages
  1983--1991, 2015.

\bibitem[Nan et~al.(2016)Nan, Wang, and Saligrama]{nan2016pruning}
F.~Nan, J.~Wang, and V.~Saligrama.
\newblock Pruning random forests for prediction on a budget.
\newblock In \emph{Advances in Neural Information Processing Systems}, pages
  2334--2342, 2016.

\bibitem[Tan(1993)]{tan1993cost}
M.~Tan.
\newblock Cost-sensitive learning of classification knowledge and its
  applications in robotics.
\newblock \emph{Machine Learning}, 13\penalty0 (1):\penalty0 7--33, 1993.

\bibitem[Trapeznikov and Saligrama(2013)]{trapeznikov2013supervised}
K.~Trapeznikov and V.~Saligrama.
\newblock Supervised sequential classification under budget constraints.
\newblock In \emph{Artificial Intelligence and Statistics}, pages 581--589,
  2013.

\bibitem[Van~Hasselt et~al.(2016)Van~Hasselt, Guez, and Silver]{van2016deep}
H.~Van~Hasselt, A.~Guez, and D.~Silver.
\newblock Deep reinforcement learning with double q-learning.
\newblock In \emph{AAAI Conference on Artificial Intelligence}, pages
  2094--2100, 2016.

\bibitem[Wang et~al.(2014{\natexlab{a}})Wang, Bolukbasi, Trapeznikov, and
  Saligrama]{wang2014model}
J.~Wang, T.~Bolukbasi, K.~Trapeznikov, and V.~Saligrama.
\newblock Model selection by linear programming.
\newblock In \emph{European Conference on Computer Vision}, pages 647--662.
  Springer, 2014{\natexlab{a}}.

\bibitem[Wang et~al.(2014{\natexlab{b}})Wang, Trapeznikov, and
  Saligrama]{wang2014lp}
J.~Wang, K.~Trapeznikov, and V.~Saligrama.
\newblock An lp for sequential learning under budgets.
\newblock In \emph{Artificial Intelligence and Statistics}, pages 987--995,
  2014{\natexlab{b}}.

\bibitem[Wang et~al.(2015)Wang, Trapeznikov, and Saligrama]{wang2015efficient}
J.~Wang, K.~Trapeznikov, and V.~Saligrama.
\newblock Efficient learning by directed acyclic graph for resource constrained
  prediction.
\newblock In \emph{Advances in Neural Information Processing Systems}, pages
  2152--2160, 2015.

\bibitem[Wang et~al.(2016)Wang, Schaul, Hessel, Hasselt, Lanctot, and
  Freitas]{wang2016dueling}
Z.~Wang, T.~Schaul, M.~Hessel, H.~Hasselt, M.~Lanctot, and N.~Freitas.
\newblock Dueling network architectures for deep reinforcement learning.
\newblock In \emph{International Conference on Machine Learning}, pages
  1995--2003, 2016.

\bibitem[Xu et~al.(2012)Xu, Weinberger, and Chapelle]{xu2012greedy}
Z.~Xu, K.~Weinberger, and O.~Chapelle.
\newblock The greedy miser: learning under test-time budgets.
\newblock In \emph{Proceedings of the 29th International Coference on
  International Conference on Machine Learning}, pages 1299--1306. Omnipress,
  2012.

\bibitem[Xu et~al.(2013)Xu, Kusner, Weinberger, and Chen]{xu2013cost}
Z.~Xu, M.~Kusner, K.~Weinberger, and M.~Chen.
\newblock Cost-sensitive tree of classifiers.
\newblock In \emph{International Conference on Machine Learning}, pages
  133--141, 2013.

\bibitem[Xu et~al.(2014)Xu, Kusner, Weinberger, Chen, and
  Chapelle]{xu2014classifier}
Z.~Xu, M.~Kusner, K.~Weinberger, M.~Chen, and O.~Chapelle.
\newblock Classifier cascades and trees for minimizing feature evaluation cost.
\newblock \emph{Journal of Machine Learning Research}, 15\penalty0
  (1):\penalty0 2113--2144, 2014.

\end{thebibliography}

\end{document}